\documentclass[11pt]{article}

\usepackage[preprint]{acl}

\usepackage{times}
\usepackage{latexsym}

\usepackage[T1]{fontenc}

\usepackage[utf8]{inputenc}

\usepackage{microtype}

\usepackage{inconsolata}

\usepackage{graphicx}
\usepackage[utf8]{inputenc} 
\usepackage[T1]{fontenc}    
\usepackage{hyperref}       
\usepackage{url}            
\usepackage{booktabs}       
\usepackage{amsfonts}       
\usepackage{nicefrac}       
\usepackage{microtype}      
\usepackage{xcolor}         
\usepackage{graphicx}
\usepackage{amsmath}
\usepackage{subcaption}
\usepackage{multirow}
\usepackage[table]{xcolor}
\usepackage{algorithmic}
\usepackage{algorithm}
\usepackage{caption}
\definecolor{BestOne}{HTML}{D6F5EE}   
\definecolor{BestTwo}{HTML}{ECFBF8}   

\definecolor{WorstOne}{HTML}{EAE2E6}  
\definecolor{WorstTwo}{HTML}{F6F1F4}  

\newcommand{\bestA}[1]{\cellcolor{BestOne}#1}
\newcommand{\bestB}[1]{\cellcolor{BestTwo}#1}

\newcommand{\worstA}[1]{\cellcolor{WorstOne}#1}
\newcommand{\worstB}[1]{\cellcolor{WorstTwo}#1}
\usepackage{wrapfig}

\title{OP-Print: Tracing Membership Imprints in On-Policy Self-Distillation}

\author{
   Qi Li  \quad Xinchao Wang\thanks{Corresponding Author} \\
  National University of Singapore \\
  \texttt{liqi@u.nus.edu} \quad \texttt{xinchao@nus.edu.sg}
}

\begin{document}
\maketitle

\begin{abstract}
On-policy distillation has been widely adopted for alleviating the distribution mismatch of off-policy distillation. Self-distillation further removes the need for an external teacher by using the student’s own outputs, enhanced with extra context or demonstrations, as supervision. This teacher-free design makes on-policy self-distillation practical, yet its privacy risks remain largely unexplored. In particular, using two coupled views of the same training example may strengthen its memorization. Motivated by this, we present the first study of membership inference attacks against on-policy self-distillation. We investigate both gray-box and black-box threat models and propose a unified attack framework, named \textbf{OP-Print}. OP-Print builds on our finding that the supervision imprint in on-policy self-distillation is sparse and structured, concentrating on uncertainty-sensitive correction sites rather than across the entire rollout. Exploiting this property, we construct attack signals tailored to the two access levels: in the gray-box setting, OP-Print directly aggregates the privileged teacher-student residual correction along the same rollout; in the black-box setting, it measures restart-based output-basin stability as an observable surrogate for the hidden correction process. Across diverse models and distillation schemes, all baseline attacks remain essentially at random guessing even under gray-box access, whereas OP-Print is broadly validated to be effective in both gray-box and black-box settings.

\end{abstract}

\section{Introduction}

On-policy distillation has recently become an important paradigm for improving model performance, as it alleviates the distribution mismatch of off-policy distillation~\cite{guo2025deepseek,song2026survey,ye2026policy}. Its self-distillation variants go one step further by removing the need for an external teacher: the distillation signal is instead derived from the student model itself, augmented with additional context or demonstrations that bootstrap its own capabilities~\cite{zhao2026self,shenfeld2026self,sang2026crisp,he2026self,yin2026policy}. This teacher-free property makes on-policy self-distillation especially attractive in practice. In particular, the model is exposed to two coupled views of the same example: a standard version without privileged information and a privileged version augmented with guidance such as a reference solution. While this design improves reasoning performance, its privacy implications have remained largely unexamined. Intuitively, processing both views of the same sample may produce a stronger and more distinctive training imprint, potentially making that sample easier to memorize than under ordinary training.

Motivated by this concern, in this work, we investigate the data privacy risks of on-policy self-distillation through the lens of membership inference attacks~\cite{shi2023detecting,li2024membership,song2019membership,li2026vid,wang2025towards}. Membership inference asks whether an adversary can determine if a given sample was used in training~\cite{shokri2017membership,carlini2022membership,yeom2018privacy}. We explore this problem under a unified gray-box and black-box threat model~\cite{hu2022membership,wu2025membership,bai2024membership}. We found this problem fundamentally nontrivial: the supervision signal in on-policy self-distillation is not globally visible, nor is it well captured by coarse output-level heuristics~\cite{zhang2024min,hu2025membership,li2026vid}. Instead, it is sparse, structured, and tied to the model's internal correction process. More importantly, our analysis shows that this signal is organized primarily by token difficulty rather than absolute decoding position. In other words, privileged correction concentrates on uncertain reasoning steps, while position alone provides only weak evidence. This makes gray-box inference feasible but already nontrivial, and makes black-box inference substantially harder, since the adversary cannot observe the underlying uncertainty or correction process and only has access to the final outputs.

To address this problem, we propose a unified framework, named \textbf{OP-Print}, that targets the same hidden supervision imprint at two levels of adversary observability. Our key insight is that on-policy self-distillation leaves a sparse but structured \textit{correction imprint} concentrated on uncertainty-sensitive reasoning tokens. In the gray-box setting, this signature can be measured directly by comparing student-conditioned and privileged teacher-conditioned behaviors along the same rollout. In the black-box setting, the correction process itself is hidden, but it still leaves \textit{observable traces in the geometry of the output basin}: when generation is restarted from partial prefixes, member and non-member examples differ in how stably their continuations remain within a local solution basin. Based on this observation, OP-Print instantiates two complementary attacks. Its gray-box variant aggregates teacher-student residual discrepancy along self-rollouts into a direct membership score, while its black-box variant probes restart-based output-basin stability as a natural surrogate for the inaccessible correction signal. Together, they offer a unified view of privacy leakage: one captures the hidden correction process directly, while the other reveals its behavioral effects from outputs alone.

\begin{figure}[t]
    \centering
    \begin{subfigure}[t]{0.44\linewidth}
        \centering
        \includegraphics[width=\linewidth]{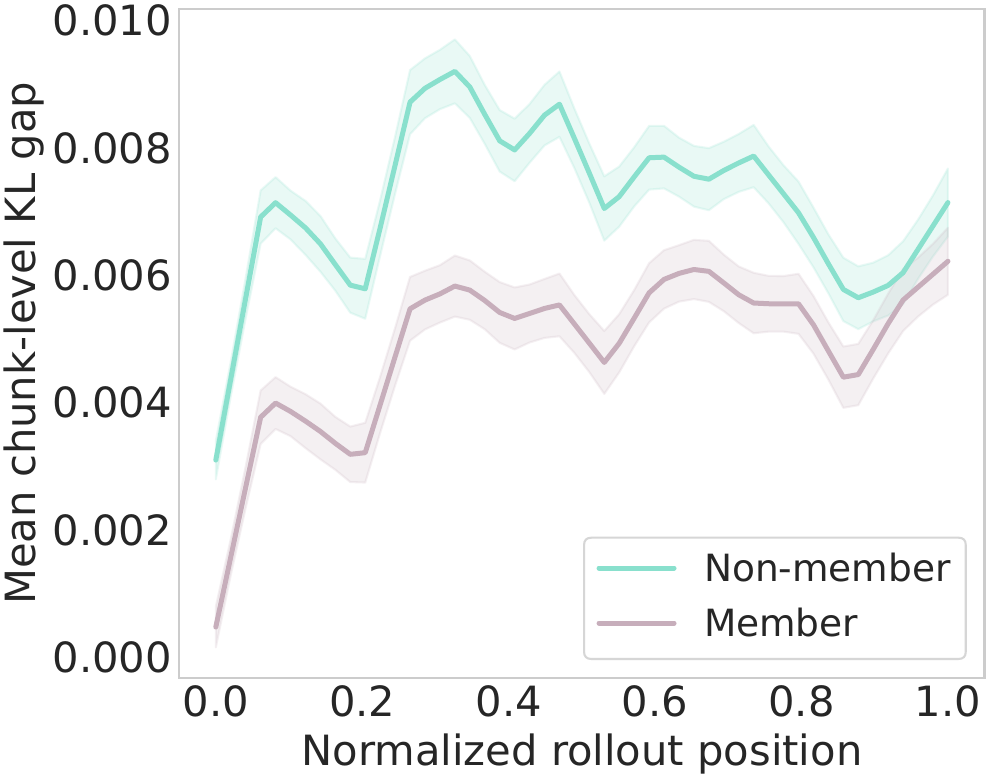}
        \caption{Gray-box Trajectory.}
        \label{fig:graybox_gap_traj}
    \end{subfigure}
    \hfill
    \begin{subfigure}[t]{0.54\linewidth}
        \centering
        \includegraphics[width=\linewidth]{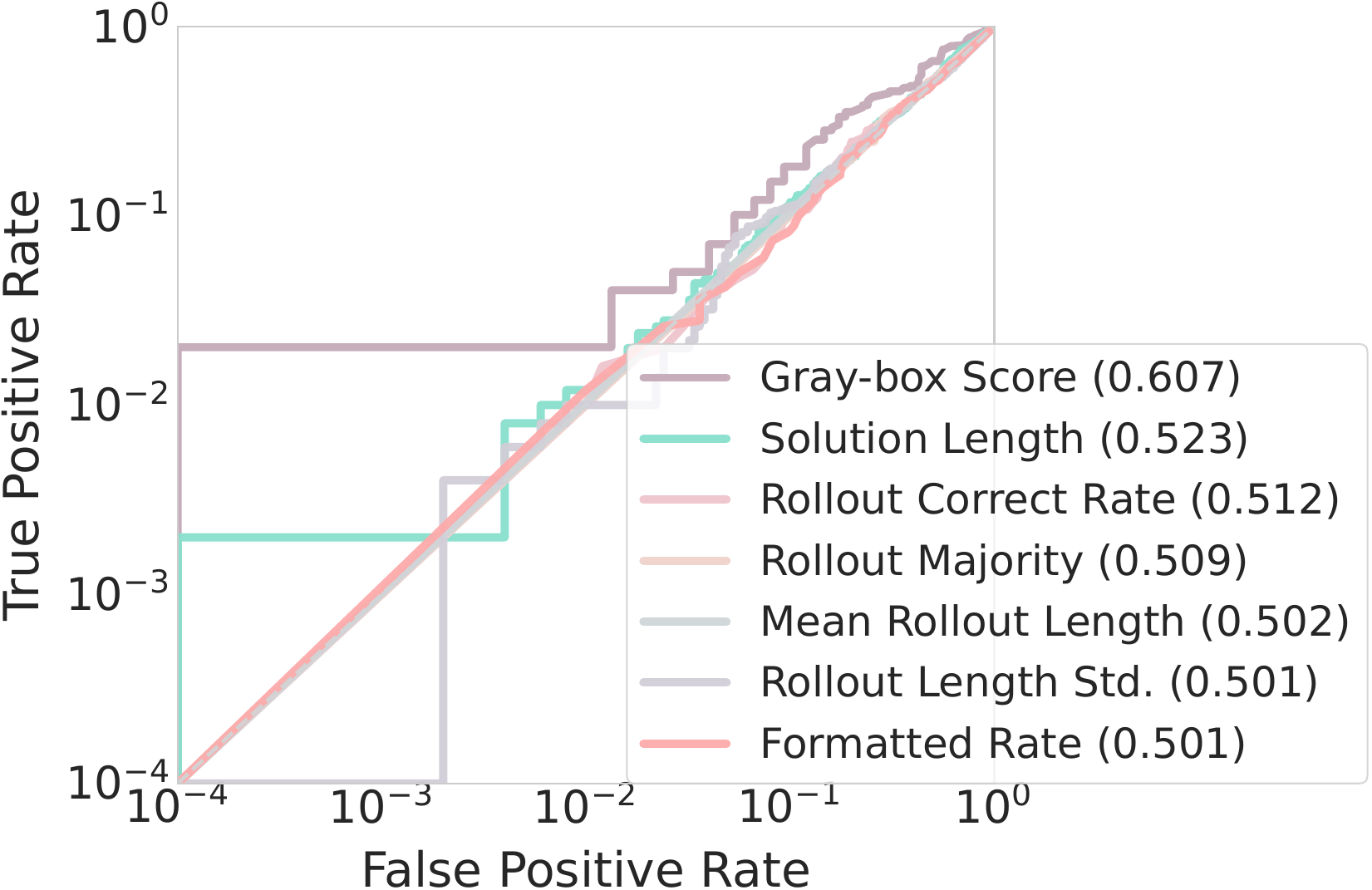}
        \caption{TPR@Low FPR.}
        \label{fig:graybox_low_fpr}
    \end{subfigure}
    \caption{
    Gray-box MIA signals from rollout-level dynamics and attack behavior (AUC) of Gray-box score.
    }
    \label{fig:graybox_wrap}
    \vspace{-4mm}
\end{figure}

A gray-box example is shown in Figure~\ref{fig:graybox_wrap}. The gap between members and non-members is observable but with magnitude vary across decoding positions. OP-Print makes use of such signal and outperforms the baselines. Across diverse models and distillation schemes, we found that all baseline attacks, consistent with observations from prior works~\cite{meeus2025sok,das2025blind,duan2024membership}, remain essentially at random guessing even under gray-box access, whereas OP-Print is broadly effective in both gray-box and black-box settings. These results reveal an overlooked privacy risk of on-policy self-distillation and show that its leakage surface differs from conventional distillation. Our contributions are as follows:

\begin{itemize}
    \item We uncover the location of the membership signal in on-policy self-distillation: rather than being globally visible, it resides in a sparse and structured \textit{correction imprint} concentrated on uncertainty-sensitive tokens.
    
    \item Building on this observation, we propose \textbf{OP-Print}, a unified attack framework that targets this hidden signal under both threat models: in the gray-box setting, it exploits privileged residual discrepancy directly, while in the black-box setting, it captures its restart-based output-basin traces, thereby enabling reliable membership inference.
    
    \item Extensive experiments across diverse LLMs consistently demonstrate the superiority of OP-Print. For example, on Qwen3-8B~\cite{yang2025qwen3} under the black-box setting, OP-Print improves AUC by 7.7\% and TPR@5\% FPR by 26.5\% over the strongest baseline.
\end{itemize}

\begin{figure*}[t]
    \centering
    \begin{subfigure}[t]{0.32\textwidth}
        \centering
        \includegraphics[width=\linewidth]{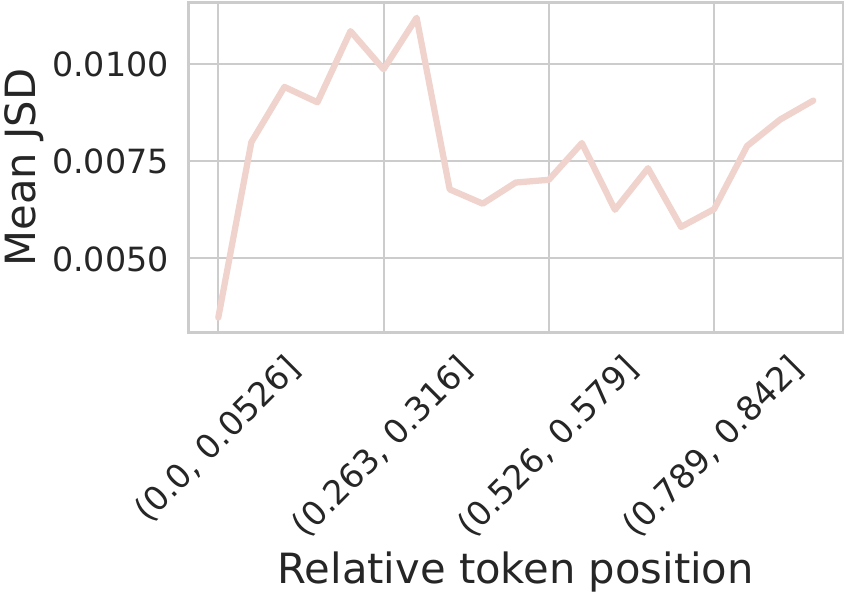}
        \caption{Mean divergence across relative token positions.}
        \label{fig:opsd_pos}
    \end{subfigure}
    \hfill
    \begin{subfigure}[t]{0.32\textwidth}
        \centering
        \includegraphics[width=\linewidth]{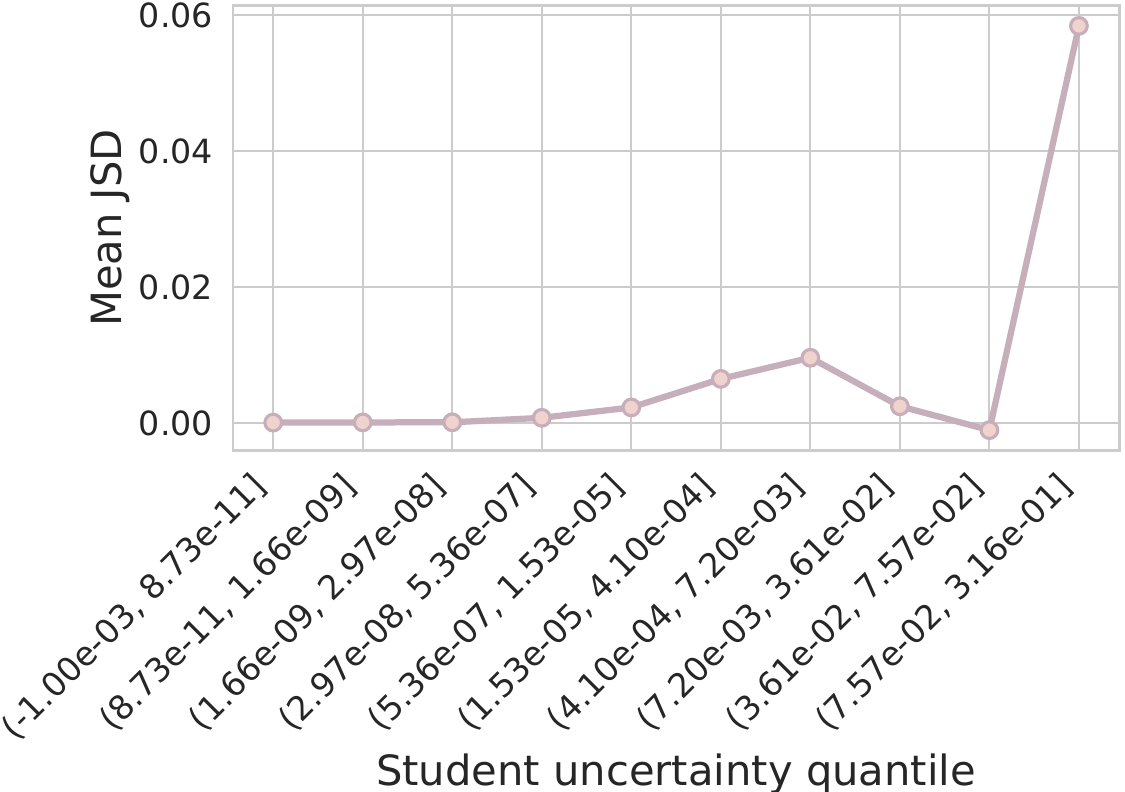}
        \caption{Mean divergence across student-uncertainty quantiles.}
        \label{fig:opsd_diff}
    \end{subfigure}
    \begin{subfigure}[t]{0.32\textwidth}
        \centering
        \includegraphics[width=\linewidth]{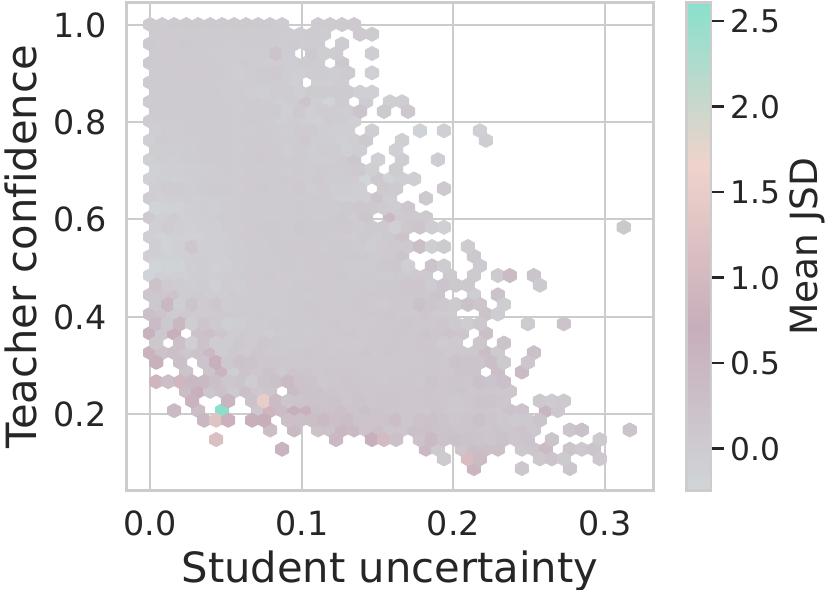}
        \caption{Reliability map of token-level teacher-student divergence.}
        \label{fig:opsd_relmap}
    \end{subfigure}
    \hfill
    \vspace{-2mm}
    \caption{
\textbf{(a).} Divergence varies only mildly across token positions, indicating that position alone is not the main source of the signal.  
\textbf{(b).} Divergence increases with student uncertainty, showing that correction concentrates on difficult reasoning tokens.  
\textbf{(c).} The reliability map reveals localized rather than uniform divergence, confirming that the supervision imprint is sparse and structured.
    }
    \vspace{-5mm}
    \label{fig:opsd_mia_mechanism}
\end{figure*}
\section{Related Work}

\noindent\textbf{On-Policy Self-Distillation.}
On-policy distillation (OPD) trains the student on trajectories sampled from its own evolving policy, reducing the training-inference mismatch of off-policy distillation~\cite{song2026survey,zhao2026self}. Existing approaches include logit matching~\cite{agarwal2024policy,ko2024distillm,jung2025todi,wu2025rethinking}, outcome supervision~\cite{ye2025black,jiang2023lion,xiong2026ovd,singh2025orpo,mendes2026didactic}, and self-distillation~\cite{zhao2026self,penaloza2026privileged,ye2026policy,ye2026online2,dat2026dopsd}. In self-distillation, the same model acts as both teacher and student under asymmetric conditioning, eliminating the need for an external teacher. Early methods such as SPIN~\cite{chen2024self} rely on reference data, whereas recent approaches inject privileged information into the teacher context. A representative example is OPSD~\cite{zhao2026self}, where the teacher receives additional supervision while the student sees only the original input, and training minimizes token-level divergence on the student’s own rollouts. Recent extensions include GATES~\cite{stein2026gates}, SDPO~\cite{hubotter2026reinforcement}, OPSDC~\cite{sang2026crisp}, and SDFT~\cite{shenfeld2026self}.

\noindent\textbf{Membership Inference Attack. }Membership Inference Attacks (MIAs)~\cite{shokri2017membership,carlini2022membership,li2026cola,yin2026refinement} determine whether a sample was used to train a target model, posing risks to sensitive training data. Existing MIAs mainly fall into metric-based and shadow model-based methods~\cite{carlini2022membership,li2024membership}. Metric-based attacks infer membership from statistics derived from model outputs~\cite{yeom2018privacy,salem2018ml,song2021systematic,li2024membership}, whereas shadow model-based attacks train auxiliary models to mimic the target~\cite{shokri2017membership,zarifzadeh2023low}. Since the latter is computationally expensive and often impractical for LLMs~\cite{li2024membership,li2026vid}, this work focuses exclusively on metric-based methods.

MIAs were first studied for classification models~\cite{shokri2017membership} and later extended to generative~\cite{chen2020gan,hayes2017logan} and embedding models~\cite{mahloujifar2021membership,song2020information}. Recent work has further investigated MIAs against LLMs~\cite{shi2023detecting,li2024membership}, covering both pretraining~\cite{shi2023detecting,zhang2024min,changpinyo2021conceptual} and post-training data~\cite{li2024membership,wu2025membership,hu2025membership}. The on-policy distillation process is commonly conducted during the post-training process. This highlights the unique importance of this problem: The capability of LLMs to effectively interact with humans fundamentally depends on the post-training stage, as the strength of this capability is directly tied to the quality of the post-training dataset. Furthermore, developers often construct their own task-specific datasets for this stage~\cite{song2026survey,zhao2026self}, which introduces additional privacy risks. Motivated by these factors, we focus on the post-training stage MIA of LLMs.

\section{Method}
\noindent\textbf{Challenges. }Figure~\ref{fig:opsd_mia_mechanism} clarifies both the difficulty and the feasibility of MIA on on-policy self-distillation. The key observation is that \textit{the supervision imprint is not organized primarily by token position, but by token difficulty.} As shown in Figure~\ref{fig:opsd_pos}, teacher-student divergence varies only moderately across relative decoding positions, so position alone provides at most a weak signal. In contrast, Figure~\ref{fig:opsd_diff} shows a much sharper dependence on student uncertainty: the divergence rises substantially in higher-uncertainty regions, indicating that privileged correction is concentrated on difficult reasoning tokens rather than uniformly distributed throughout the rollout. Figure~\ref{fig:opsd_relmap} further shows that this signal occupies only a restricted region of the reliability map, confirming that the resulting imprint is sparse and structured instead of globally exposed. This structure directly explains the gap between gray-box and black-box attacks. In the gray-box setting, adversaries can access the teacher-conditioned residual correction signal along the same rollout, and directly aggregates this privileged discrepancy. In the black-box setting, however, the most informative variable is fundamentally hidden, and the adversary only observes final output tokens. Since position is only weakly informative, this makes black-box MIA substantially harder.

\begin{figure*}[t]
    \centering
    \includegraphics[width=\textwidth]{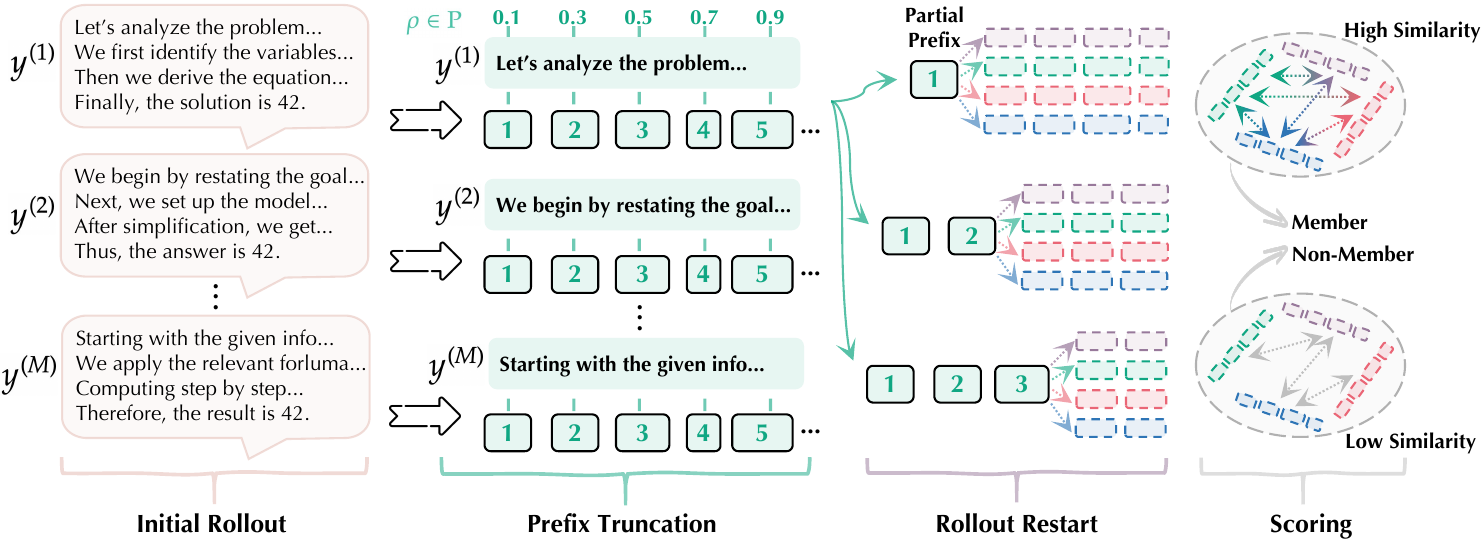} 
    \caption{An illustration of the black-box attack process of OP-Print. Given multiple initial rollouts, OP-Print truncates each rollout at different prefix ratios, restarts generation from the partial prefix, and scores membership by measuring the similarity between restarted outputs.
Members tend to produce more stable and similar restart trajectories, while non-members exhibit lower restart consistency.
}
    \vspace{-3mm}
    \label{fig:pipeline}
\end{figure*}
\noindent\textbf{Problem Setup. }Let $x=(q,r)$ denote a candidate training example, where $q$ is the problem and $r$ is its reference solution. 
Given a target checkpoint $p_{\theta}$, our goal is to infer whether $x$ belongs to the member set used during self-distillation.
We write $p_s(q)$ for the standard student prompt constructed from $q$ alone, and $p_t(q,r)$ for the privileged prompt that additionally conditions on the reference solution $r$.
For a rollout $y=(y_1,\dots,y_T)\sim p_{\theta}(\cdot\mid p_s(q))$, let $y_{<t}$ denote its prefix before token $t$.

\noindent\textbf{A Unified View: Hidden Correction Imprint. }The analysis in Figure~\ref{fig:opsd_mia_mechanism} suggests that on-policy self-distillation leaves a \emph{structured correction imprint}: membership-relevant signal is sparse, concentrated on uncertainty-sensitive reasoning tokens, and only weakly aligned with absolute decoding position.
This motivates a unified view of both attacks.
We posit a latent quantity measuring how much privileged correction the model would still require on a rollout:
\begin{equation}
\begin{aligned}
\Delta_t(x,y)
&=
\mathrm{KL}\!\left(
p_{\theta}(\cdot \mid p_t(q,r), y_{<t})
\right.\\
&\qquad\left\|
p_{\theta}(\cdot \mid p_s(q), y_{<t})
\right).
\end{aligned}
\label{eq:token_gap}
\end{equation}
Intuitively, $\Delta_t(x,y)$ is large when the student-conditioned model still disagrees substantially with its privileged counterpart at token $t$, and small when the correction has already been internalized. Members should therefore exhibit \emph{smaller} privileged correction on average, because self-distillation has already adapted the model on those examples.
Our gray-box attack estimates this quantity directly, while our black-box attack targets its observable consequence in the output basin.
\begin{table}[t]
\centering
\small
\hrule
\vspace{0.4em}
\textbf{Algorithm 1: OP-Print (Black-box)}
\vspace{0.2em}
\hrule
\vspace{0.4em}

\begin{algorithmic}[1]
\REQUIRE Target model $p_\theta$, query $q$, prefix ratios $\mathcal{P}$,
number of full generations $M$, number of restarts $R$
\ENSURE Membership score $S_{\mathrm{black}}(x)$

\STATE Sample $M$ generations
$\{y^{(m)}\}_{m=1}^M \sim p_\theta(\cdot \mid p_s(q))$

\FOR{$m=1,\dots,M$}
    \FOR{each $\rho \in \mathcal{P}$}
        \STATE Truncate $y^{(m)}$ into prefix $\pi_\rho(y^{(m)})$
        \STATE Restart decoding $R$ times from $\pi_\rho(y^{(m)})$

        \STATE Obtain restart set $\mathcal{R}_{m,\rho}(x)$, where\\
        \hspace*{2em}
        $\displaystyle
        \mathcal{R}_{m,\rho}(x)=
        \{\tilde y^{(m,\rho,1)},\dots,\tilde y^{(m,\rho,R)}\}.
        $

        \STATE Compute restart stability\\[-0.4em]
        \hspace*{1.2em}
        $\displaystyle
        \begin{aligned}
        \mathrm{Stab}_{\mathrm{text}}(\mathcal{R}_{m,\rho}(x))
        &=
        \frac{2}{R(R-1)}
        \sum_{i<j} \\
        &\qquad
        \mathrm{sim}\!\left(
        \tilde y^{(m,\rho,i)},
        \tilde y^{(m,\rho,j)}
        \right)
        \end{aligned}
        $
    \ENDFOR
\ENDFOR

\STATE Aggregate over all full generations and prefix ratios:\\
\hspace*{1.2em}
$\displaystyle
S_{\mathrm{black}}(x)
=
\frac{1}{M|\mathcal{P}|}
\sum_{m=1}^{M}
\sum_{\rho\in\mathcal{P}}
\mathrm{Stab}_{\mathrm{text}}(\mathcal{R}_{m,\rho}(x))
$

\RETURN $S_{\mathrm{black}}(x)$
\end{algorithmic}

\vspace{0.2em}
\hrule
\label{pseudo_code}
\end{table}

\noindent\textbf{Gray-box Attack: Privileged Residual Aggregation.}
In the gray-box setting, we can evaluate the target model under both $p_s(q)$ and $p_t(q,r)$ along the \emph{same} rollout.
Given $K$ sampled rollouts $\{y^{(k)}\}_{k=1}^K$ from $p_{\theta}(\cdot\mid p_s(q))$, we define the gray-box score as
\begin{equation}
S_{\mathrm{gray}}(x)
=
-
\frac{1}{K}
\sum_{k=1}^{K}
\frac{1}{T_k}
\sum_{t=1}^{T_k}
\Delta_t\!\left(x,y^{(k)}\right).
\label{eq:gray_score}
\end{equation}
Here, a larger score means a smaller average privileged gap, hence stronger evidence that the example has already been absorbed by the self-distilled model.
In implementation, we compute the divergence in fixed-size chunks (e.g., 64 tokens) for efficiency (ablations can be found in Section~\ref{ablation}). In our experiments, we also introduce a grey-box scoring variant to assess the utility of the maximum-discrepancy position. This variant can be written as \(
S_{\mathrm{gray}}^{\max}(x)
=
-
\frac{1}{K}
\sum_{k=1}^{K}
\max_{1 \le t \le T_k}
\Delta_t\!\left(x,y^{(k)}\right)
\).

\begin{table*}[t]
\centering
\caption{Gray-box and black-box results with the JSD objective. Colors indicate the top-2 and worst-2 results in the gray-box setting, and the top-1 and worst-1 results in the black-box setting.}
\label{tab:greybox_results}
\resizebox{\textwidth}{!}{
\begin{tabular}{lccc|ccc|ccc}
\toprule
\multirow{2}{*}{Method}
& \multicolumn{3}{c}{Qwen3-1.7B}
& \multicolumn{3}{c}{Qwen3-4B}
& \multicolumn{3}{c}{Qwen3-8B} \\
\cmidrule(lr){2-4}
\cmidrule(lr){5-7}
\cmidrule(lr){8-10}
& AUC & Acc. & TPR@5\%FPR
& AUC & Acc. & TPR@5\%FPR
& AUC & Acc. & TPR@5\%FPR \\
\midrule
\multicolumn{10}{l}{\cellcolor{gray!5}\textbf{Gray-box setting}} \\
\midrule

\textbf{Min-K\% Prob}
& 0.512 & 0.518 & 0.052
& \worstA{0.497} & \worstA{0.505} & \worstB{0.046}
& 0.529 & 0.511 & 0.055 \\

\textbf{ModR\'enyi} \textbf{($\alpha=0.5$)}
& \worstA{0.484} & 0.506 & \worstA{0.044}
& 0.508 & 0.508 & 0.051
& 0.513 & 0.514 & 0.056 \\

\textbf{ModR\'enyi} \textbf{($\alpha=2$)}
& 0.516 & 0.514 & 0.057
& 0.507 & 0.512 & 0.047
& 0.506 & \worstA{0.502} & 0.050 \\

\textbf{R\'enyi} \textbf{($\alpha=0.5$)}
& 0.503 & 0.509 & 0.049
& 0.517 & 0.511 & 0.054
& \worstB{0.495} & \worstB{0.503} & \worstA{0.045} \\

\textbf{R\'enyi} \textbf{($\alpha=1$)}
& 0.508 & \worstA{0.501} & \worstB{0.046}
& 0.512 & \worstB{0.507} & 0.050
& 0.507 & 0.513 & 0.056 \\

\textbf{R\'enyi} \textbf{($\alpha=2$)}
& 0.517 & 0.518 & 0.055
& \worstB{0.506} & 0.508 & \worstA{0.044}
& \worstA{0.491} & 0.503 & 0.052 \\

\textbf{R\'enyi} \textbf{($\alpha=\infty$)}
& \worstB{0.489} & \worstB{0.502} & 0.048
& 0.514 & 0.511 & 0.057
& 0.511 & 0.509 & \worstB{0.046} \\

\midrule
\textbf{OP-Print-G}
& \bestB{0.579} & \bestA{0.586} & \bestB{0.108}
& \bestB{0.580} & \bestB{0.583} & \bestB{0.095}
& \bestB{0.556} & \bestB{0.543} & \bestB{0.088} \\

\textbf{OP-Print-G-Max}
& \bestA{0.594} & \bestB{0.583} & \bestA{0.110}
& \bestA{0.611} & \bestA{0.604} & \bestA{0.127}
& \bestA{0.587} & \bestA{0.580} & \bestA{0.102} \\

\midrule
\multicolumn{10}{l}{\cellcolor{gray!5}\textbf{Black-box setting}} \\
\midrule

\textbf{S-Len}
& 0.505 & 0.516 & 0.068
& \worstA{0.419} & \worstA{0.507} & 0.060
& 0.505 & 0.516 & 0.068 \\

\textbf{RC-Rate}
& 0.507 & 0.517 & 0.034
& 0.544 & 0.566 & \bestA{0.104}
& 0.506 & 0.508 & 0.046 \\

\textbf{R-Maj}
& 0.516 & 0.523 & 0.000
& 0.450 & 0.508 & 0.000
& 0.493 & 0.523 & 0.000 \\

\textbf{Avg. R-Len}
& \worstA{0.495} & \worstA{0.508} & \worstA{0.000}
& 0.458 & 0.508 & \worstA{0.000}
& \worstA{0.489} & \worstA{0.506} & \worstA{0.000} \\

\textbf{R-Std}
& 0.536 & 0.528 & 0.066
& 0.542 & 0.568 & 0.084
& 0.511 & 0.515 & 0.046 \\

\textbf{F-Rate}
& 0.508 & 0.513 & 0.048
& 0.516 & 0.543 & 0.098
& 0.517 & 0.521 & 0.052 \\

\midrule
\textbf{OP-Print-B}
& \bestA{0.554} & \bestA{0.552} & \bestA{0.086}
& \bestA{0.575} & \bestA{0.573} & 0.089
& \bestA{0.557} & \bestA{0.546} & \bestA{0.086} \\

\bottomrule
\end{tabular}
}
\vspace{-4mm}
\end{table*}
\noindent\textbf{Black-box Attack: Restart-based Basin Stability.}
In the black-box setting, the informative variable in Figure~\ref{fig:opsd_mia_mechanism}, i.e., the teacher-conditioned residual correction, is hidden.
The adversary sees only generated text, so a direct estimate of Equation~\ref{eq:token_gap} is unavailable.
Our key design choice is therefore to attack the \emph{behavioral trace} left by hidden correction, rather than the hidden correction itself.

As illustrated in Figure~\ref{fig:pipeline}, for each input $q$, we first sample $M$ full generations $\{y^{(m)}\}_{m=1}^M$ from $p_{\theta}(\cdot\mid p_s(q))$.
For each generation $y^{(m)}$ and each prefix ratio $\rho\in\mathcal{P}$, we truncate it to a partial prefix $\pi_{\rho}(y^{(m)})$ and restart decoding from that prefix multiple times, obtaining a restart set
\begin{equation}
\mathcal{R}_{m,\rho}(x)
=
\left\{
\tilde y^{(m,\rho,1)},\dots,\tilde y^{(m,\rho,R)}
\right\}.
\end{equation}
We then measure how tightly these restarted continuations collapse in text space:
\begin{equation}
\begin{aligned}
\mathrm{Stab}_{\mathrm{text}}\!\left(\mathcal{R}_{m,\rho}(x)\right)
&=
\frac{2}{R(R-1)}
\sum_{i<j} \\
&\quad
\mathrm{sim}\!\left(
\tilde y^{(m,\rho,i)},
\tilde y^{(m,\rho,j)}
\right),
\end{aligned}
\label{eq:text_stability}
\end{equation}
where $\mathrm{sim}(\cdot,\cdot)$ is a normalized string-level similarity.
Our final black-box score is the average restart-text stability across sampled full generations and prefix ratios:
\begin{equation}
S_{\mathrm{black}}(x)
=
\frac{1}{M|\mathcal{P}|}
\sum_{m=1}^{M}
\sum_{\rho\in\mathcal{P}}
\mathrm{Stab}_{\mathrm{text}}\!\left(\mathcal{R}_{m,\rho}(x)\right).
\label{eq:black_score}
\end{equation}
A higher $S_{\mathrm{black}}(x)$ indicates a narrower and more stable output basin under prefix restarts, which we interpret as evidence that the hidden correction has already been more strongly internalized, and hence that $x$ is more likely to be a member. The pseudo code is provided in Algorithm 1.

\begin{figure*}[ht]
    \centering
    \makebox[0.7\textwidth][c]{
        \begin{subfigure}[t]{0.34\textwidth}
            \centering
            \includegraphics[width=\linewidth]{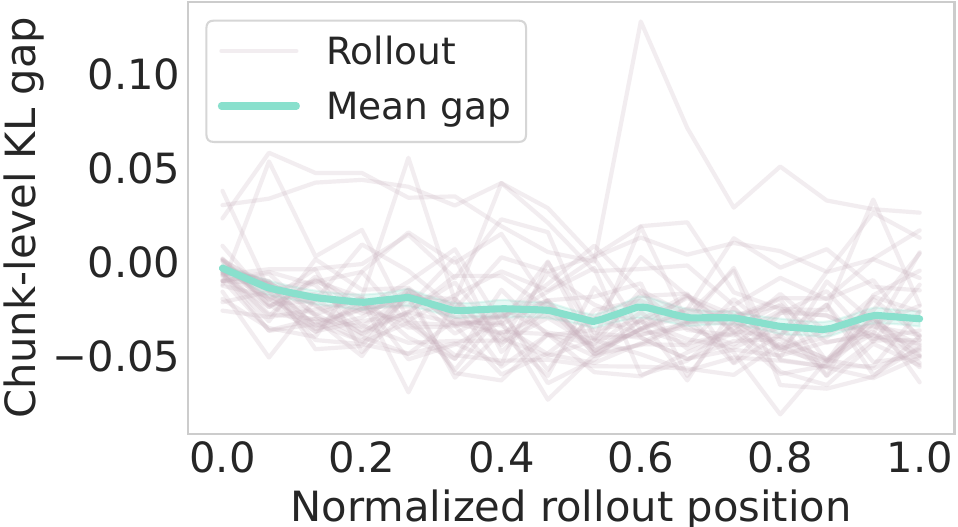}
            \caption{Gray-box member case.}
            \label{fig:opsd_gray_member_case}
        \end{subfigure}
        \hfill
        \begin{subfigure}[t]{0.34\textwidth}
            \centering
            \includegraphics[width=\linewidth]{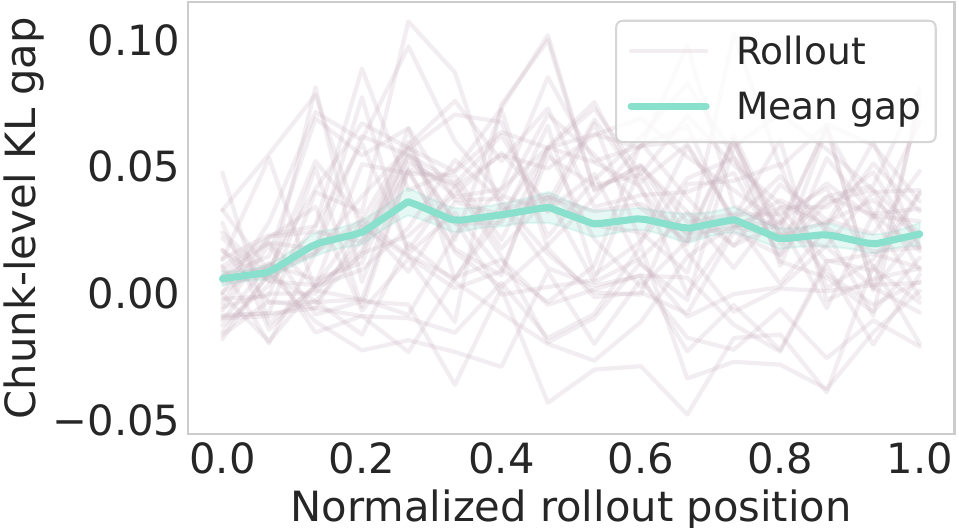}
            \caption{Gray-box non-member case.}
            \label{fig:opsd_gray_nonmember_case}
        \end{subfigure}
    }

    \vspace{0.6em}

    \begin{subfigure}[t]{0.48\textwidth}
        \centering
        \includegraphics[width=\linewidth]{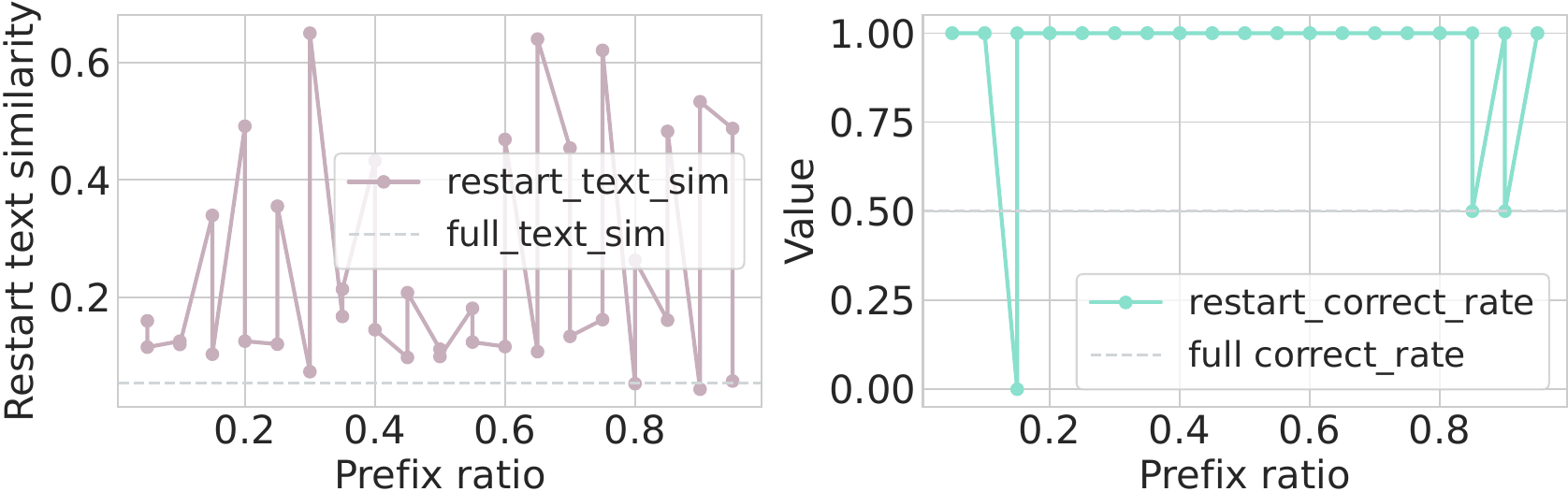}
        \caption{Black-box member case.}
        \label{fig:opsd_black_member_case}
    \end{subfigure}
    \hfill
    \begin{subfigure}[t]{0.48\textwidth}
        \centering
        \includegraphics[width=\linewidth]{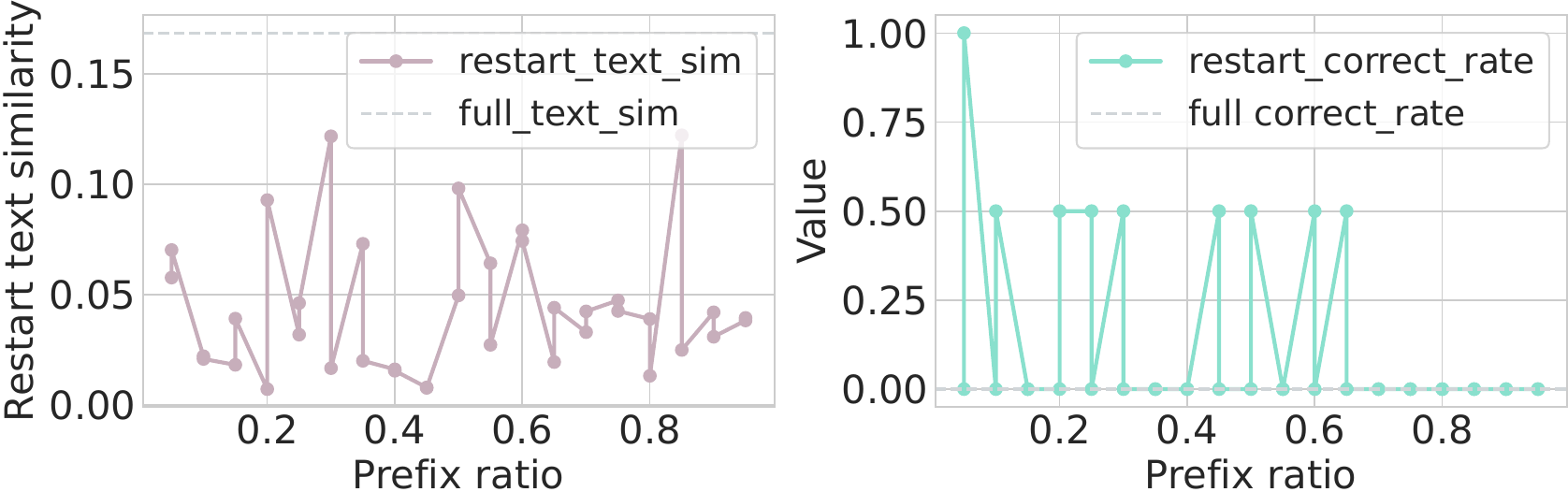}
        \caption{Black-box non-member case.}
        \label{fig:opsd_black_nonmember_case}
    \end{subfigure}
    \vspace{-1mm}
    \caption{
Representative gray-box and black-box OP-Print signals. \textbf{(a)(b)} In the gray-box setting, members show lower chunk-level KL gaps and higher scores than non-members. \textbf{(c)(d)} In the black-box setting, members exhibit more stable outputs under prefix restarts, whereas non-members are less stable.
    }
    \label{fig:opsd_mia_case_studies}
    \vspace{-4mm}
\end{figure*}
\section{Experiments}

\noindent\textbf{Distillation Variants.}
We evaluate our attacks on two representative on-policy self-distillation variants. Both follow the same high-level training paradigm~\cite{zhao2026self,song2026survey}: the student first generates an on-policy completion from the original problem, and the resulting continuation is then paired with two conditioning contexts with teacher-side prompt contains additional privileged information derived from the reference solution. In the first variant, we follow the practice of previous works~\cite{zhao2026self,shenfeld2026self} to use a generalized Jensen-Shannon divergence (JSD) objective that aligns the teacher and student distributions over the student-generated continuation. For the second variant, we follow a tinker-style~\cite{lu2025onpolicydistillation} formulation. It preserves the same privileged teacher / standard student setup, but replaces distribution matching with a sampled-token policy-improvement objective. Concretely, the teacher-student log-probability difference on the generated continuation is used as an advantage-like signal, and optimization increases the student probability on tokens preferred by the privileged teacher. This yields a more policy-oriented realization of on-policy self-distillation. These two variants provide a useful contrast between distribution-level and sampled-token-level self-distillation under the same on-policy training principle. All training hyperparameters are deferred to Appendix~\ref{train_detail}.

\noindent\textbf{Datasets and Models.}
A long-standing and highly debated issue in MIA against LLMs is data construction~\cite{meeus2025sok,das2025blind,duan2024membership}. In prior work, the attack results are often rendered unreliable by severe distribution mismatch between the member and non-member sets, leading to overclaimed performance~\cite{meeus2025sok}. On the other side, prior work has found that training a model too much on-policy self-distillation can suffer from mode collapse~\cite{song2026survey,zhao2026self}. This also motivates controlling the data scale in evaluation.
To avoid the above problems and to ensure a correct and fair evaluation of attack performance, we carefully select and construct the dataset and train the target models under a protocol that strictly follows the requirements of membership inference evaluation. For the dataset, e use the mathematical reasoning subset of OpenThoughts~\cite{guha2025openthoughts}, sampling 1k problem-solution pairs with chain-of-thought reasoning, splitting them evenly into 500 members and 500 non-members. For the target models, we fine-tune three Qwen3 variants~\cite{yang2025qwen3} (1.7B, 4B, 8B) on the sampled member set under the distillation variants described above. Training details are in Appenfix~\ref{train_detail}.

\noindent\textbf{Baselines.}
We consider both the general gray-box baselines and on-policy self-distillation specified black-box baselines. For gray-box baselines, we include Min-$K\%$~\cite{zhang2024min} and MaxRényi-$K\%$~\cite{li2024membership}. We report the best results among different variants of these methods.
For black-box baselines, we construct rollout-derived metrics including solution length (S-Len), rollout correct-rate (RC-Rate), rollout majority-answer (R-Maj), average rollout length (Avg. R-Len), rollout-length standard deviation (R-Std), and formatted-answer rate (F-Rate). These metrics capture output length, correctness, answer consistency, length stability, and formatting reliability, respectively.

\begin{figure*}[t]
    \centering
    \begin{subfigure}[t]{0.50\textwidth}
        \centering
        \includegraphics[width=\linewidth]{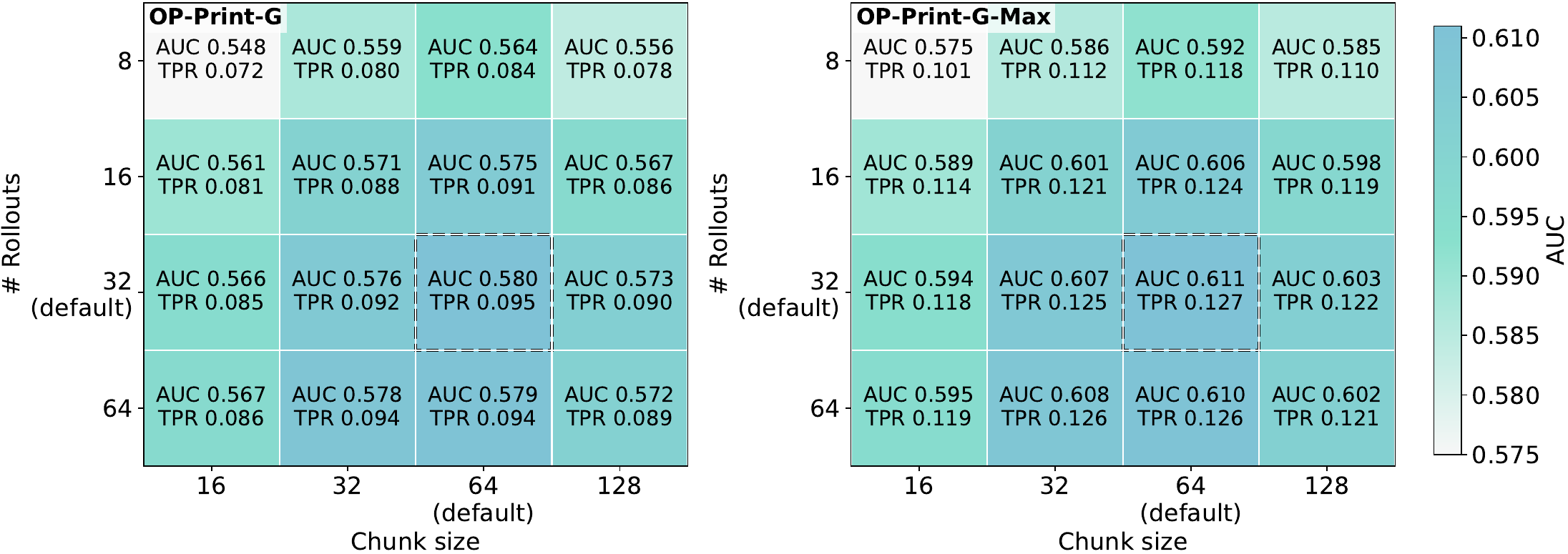}
        \caption{Gray-box sensitivity to the \# rollouts and the chunk size. The default setting is outlined with a dashed box.}
        \label{fig:ablation_graybox_heatmap}
    \end{subfigure}
    \hfill
    \begin{subfigure}[t]{0.24\textwidth}
        \centering
        \includegraphics[width=\linewidth]{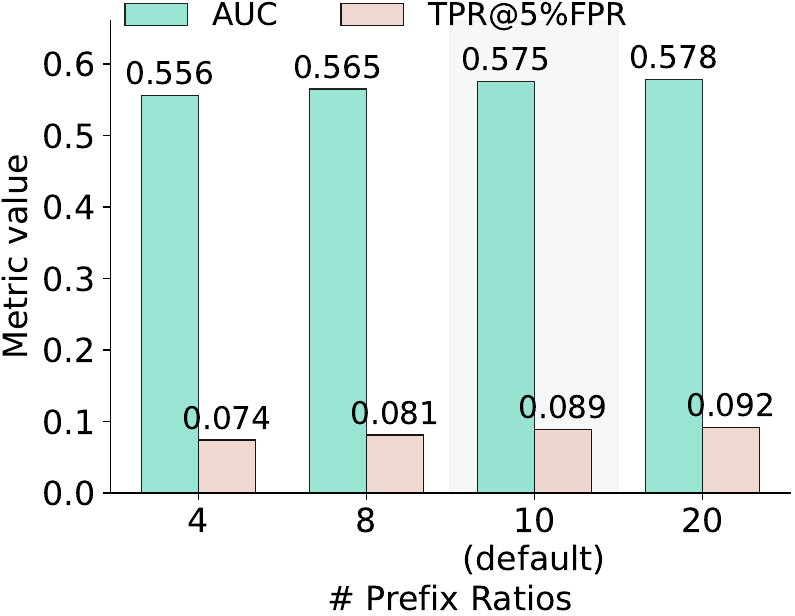}
        \caption{Black-box sensitivity to the number of prefix ratios.}
        \label{fig:ablation_blackbox_prefix}
    \end{subfigure}
    \hfill
    \begin{subfigure}[t]{0.20\textwidth}
        \centering
        \includegraphics[width=\linewidth]{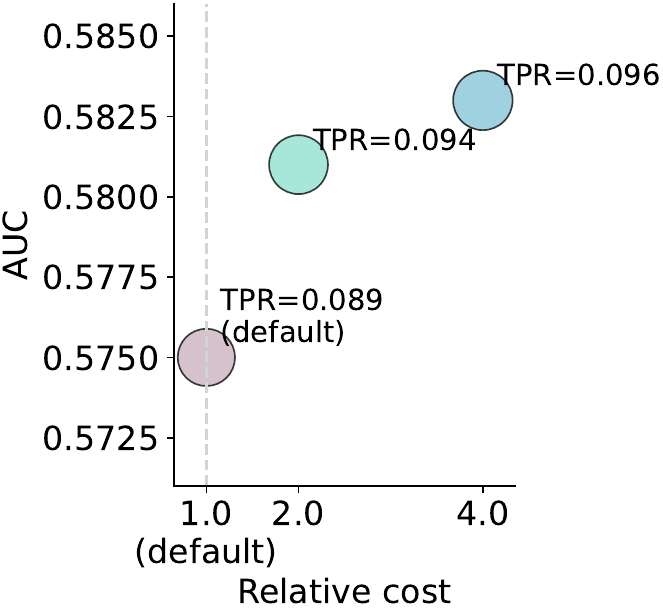}
        \caption{Trade-off of restart cost and performance.}
        \label{fig:ablation_restart_tradeoff}
    \end{subfigure}

    \caption{
    Ablation studies of OP-Print on Qwen3-4B. \textbf{(a)} Gray-box performance across rollout numbers and chunk sizes. \textbf{(b)} Black-box performance with varying numbers of prefix ratios. \textbf{(c)} Restart cost-performance trade-off, where bubble size denotes TPR@5\%FPR.
    }
    \label{fig:opprint_ablation}
    \vspace{-3mm}
\end{figure*}
\noindent\textbf{Evaluation metric. }As a binary classification problem, the performance can be evaluated with the AUC score~\cite{carlini2021extracting}. We define the members as "positive" and the non-members as "negative". We also report True Positive Rate (TPR) at low False Positive Rate (FPR)~\cite{carlini2022membership}, which is an important metric in MIAs and measures detection rate at a meaningful threshold. We set the threshold as 5\% and evaluate all the methods under TPR@5\% FPR. We also report the best classification accuracy achievable by sweeping over all possible thresholds on the attack scores. Specifically, this accuracy is computed as the maximum value of $1 - \frac{\text{FPR} + (1 - \text{TPR})}{2}$ across the ROC curve, representing the optimal classification accuracy between members and non-members. We use \# frames to denote the number of frames.

\subsection{Main Results}
\noindent\textbf{Results and Analysis.}
Table~\ref{tab:greybox_results} reports membership inference results under both gray-box and black-box settings. For gray-box attacks, we set the chunk size to 64 and $K=32$; for black-box attacks, we use $|\mathcal{P}|=10$ with $\rho\sim\mathrm{Uniform}(0,1)$ (see Section~\ref{ablation} for ablations). Distillation uses the dense token-level JSD objective. Overall, OP-Print consistently outperforms existing baselines. In the gray-box setting, OP-Print-G-Max achieves the highest AUC across all model scales, improving the strongest non-OP-Print baseline from 0.517 to 0.594 on Qwen3-1.7B, 0.517 to 0.611 on Qwen3-4B, and 0.529 to 0.587 on Qwen3-8B. It also achieves TPR@5\%FPR scores of 0.110, 0.127, and 0.102, respectively, demonstrating strong performance in the low-FPR regime.

Standard likelihood- and entropy-based MIA baselines, including Min-K\% Prob and R\'enyi scores, remain near random guessing, with AUCs around 0.49--0.52. In contrast, OP-Print-G and OP-Print-G-Max perform substantially better, indicating that on-policy self-distillation leaves a structured correction imprint that is better captured by teacher--student discrepancy than by raw likelihood statistics. The consistent advantage of OP-Print-G-Max further suggests that membership information is concentrated at the most discrepant positions.

In the black-box setting, OP-Print-B achieves the best overall AUC and accuracy across model scales. For example, on Qwen3-4B, it reaches 0.575 AUC and 0.573 accuracy, outperforming all black-box baselines on both metrics. Although RC-Rate attains the highest TPR@5\%FPR on Qwen3-4B, its gains are less consistent across scales. These results suggest that black-box membership evidence is better captured by aggregating behavioral consequences of the hidden correction imprint than by any single surface heuristic.

\begin{table*}[t]
\centering
\caption{Gray-box and black-box results with tinker-style objective. Colors indicate the top-2 and worst-2 results in the gray-box setting, and the top-1 and worst-1 results in the black-box setting.}
\label{tab:greybox_results_tinker}
\resizebox{\textwidth}{!}{
\begin{tabular}{lccc|ccc|ccc}
\toprule
\multirow{2}{*}{Method}
& \multicolumn{3}{c}{Qwen3-1.7B}
& \multicolumn{3}{c}{Qwen3-4B}
& \multicolumn{3}{c}{Qwen3-8B} \\
\cmidrule(lr){2-4}
\cmidrule(lr){5-7}
\cmidrule(lr){8-10}
& AUC & Acc. & TPR@5\%FPR
& AUC & Acc. & TPR@5\%FPR
& AUC & Acc. & TPR@5\%FPR \\
\midrule
\multicolumn{10}{l}{\cellcolor{gray!5}\textbf{Gray-box setting}} \\
\midrule

\textbf{Min-K\% Prob}
& 0.509 & 0.510 & 0.051
& \worstA{0.491} & \worstA{0.502} & \worstB{0.045}
& 0.522 & 0.503 & 0.054 \\

\textbf{ModR\'enyi} \textbf{($\alpha=0.5$)}
& \worstA{0.478} & \worstB{0.504} & \worstA{0.042}
& 0.503 & 0.500 & 0.052
& 0.506 & 0.507 & 0.055 \\

\textbf{ModR\'enyi} \textbf{($\alpha=2$)}
& 0.507 & 0.505 & 0.055
& \worstB{0.496} & 0.508 & 0.046
& 0.500 & \worstA{0.501} & 0.049 \\

\textbf{R\'enyi} \textbf{($\alpha=0.5$)}
& 0.495 & 0.502 & 0.050
& 0.510 & 0.504 & 0.056
& \worstB{0.484} & \worstB{0.500} & \worstA{0.043} \\

\textbf{R\'enyi} \textbf{($\alpha=1$)}
& 0.500 & 0.500 & \worstB{0.045}
& 0.504 & 0.503 & 0.048
& 0.502 & 0.505 & 0.057 \\

\textbf{R\'enyi} \textbf{($\alpha=2$)}
& 0.505 & 0.512 & 0.053
& 0.498 & \worstB{0.501} & \worstA{0.042}
& \worstA{0.481} & 0.500 & 0.051 \\

\textbf{R\'enyi} \textbf{($\alpha=\infty$)}
& \worstB{0.481} & \worstA{0.500} & 0.047
& 0.506 & 0.509 & 0.054
& 0.506 & 0.504 & \worstB{0.048} \\

\midrule
\textbf{OP-Print-G}
& \bestB{0.571} & \bestA{0.579} & \bestB{0.076}
& \bestB{0.563} & \bestA{0.575} & \bestA{0.094}
& \bestB{0.548} & \bestB{0.553} & \bestB{0.071} \\

\textbf{OP-Print-G-Max}
& \bestA{0.583} & \bestB{0.578} & \bestA{0.109}
& \bestA{0.571} & \bestB{0.569} & \bestB{0.083}
& \bestA{0.579} & \bestA{0.576} & \bestA{0.098} \\

\midrule
\multicolumn{10}{l}{\cellcolor{gray!5}\textbf{Black-box setting}} \\
\midrule

\textbf{S-Len}
& 0.497 & 0.506 & 0.067
& \worstA{0.481} & 0.500 & 0.058
& 0.497 & 0.507 & 0.066 \\

\textbf{RC-Rate}
& 0.500 & 0.510 & 0.033
& 0.533 & 0.534 & 0.082
& 0.500 & 0.505 & 0.044 \\

\textbf{R-Maj}
& 0.507 & 0.514 & 0.000
& 0.489 & \worstA{0.500} & 0.000
& 0.512 & 0.504 & 0.000 \\

\textbf{Avg. R-Len}
& \worstA{0.497} & \worstA{0.501} & \worstA{0.000}
& 0.487 & 0.500 & \worstA{0.000}
& \worstA{0.480} & \worstA{0.502} & \worstA{0.000} \\

\textbf{R-Std}
& 0.527 & 0.520 & 0.064
& 0.539 & 0.540 & \bestA{0.093}
& 0.505 & 0.507 & 0.044 \\

\textbf{F-Rate}
& 0.503 & 0.502 & 0.046
& 0.508 & 0.534 & 0.096
& 0.530 & 0.532 & \bestA{0.072} \\

\midrule
\textbf{OP-Print-B}
& \bestA{0.548} & \bestA{0.550} & \bestA{0.085}
& \bestA{0.567} & \bestA{0.562} & 0.091
& \bestA{0.553} & \bestA{0.539} & 0.064 \\

\bottomrule
\end{tabular}
}
\vspace{-3mm}
\end{table*}

Across scales, the results also show that membership leakage under on-policy self-distillation does not monotonically vanish as the backbone becomes larger. Both gray-box and black-box OP-Print variants remain effective on Qwen3-8B, demonstrating that the proposed attack captures a persistent training-induced signal rather than a scale-specific artifact. These findings validate our central claim: on-policy self-distillation introduces a distinct membership leakage channel, and exploiting the correction imprint yields stronger and more reliable inference than conventional MIA scores or simple rollout statistics.

\noindent\textbf{Case-level Interpretation.}
Figure~\ref{fig:opsd_mia_case_studies} illustrates this correspondence qualitatively.
In the gray-box cases ($K=32$), the member example exhibits a lower chunk-level KL gap throughout the rollout and therefore a higher $S_{\mathrm{gray}}$, while the non-member case shows the opposite pattern.
In the black-box cases ($|\mathcal{P}|=20$, $\rho\sim \mathrm{Uniform}(0,1)$), the member example produces a visibly more stable restart basin, whereas the non-member case is more dispersive under the same restart procedure.
These examples are consistent with our design principle: gray-box MIA accesses privileged correction directly, while black-box MIA recovers its observable traces through restart-based output stability.

\noindent\textbf{Influence of Distillation Variants. }Table~\ref{tab:greybox_results_tinker} shows that OP-Print remains effective under the tinker-style objective, indicating that the correction-imprint signal generalizes beyond a specific distillation formulation. In the gray-box setting, OP-Print-G and OP-Print-G-Max consistently outperform likelihood-based baselines. OP-Print-G-Max achieves the best AUC on Qwen3-1.7B and Qwen3-8B, while OP-Print-G remains competitive on several metrics, further supporting that membership information is concentrated at highly informative discrepancy positions.
In the black-box setting, OP-Print-B remains the strongest overall method in AUC and accuracy across model scales. While simple rollout heuristics such as R-Std and F-Rate occasionally achieve higher TPR@5\%FPR on specific models, their performance is less consistent. In contrast, OP-Print-B delivers more stable gains, showing that the behavioral footprint of the hidden correction imprint remains observable even without privileged gray-box signals.

\subsection{Ablation Study}
\label{ablation}
We further analyze the sensitivity of OP-Print to key hyperparameters, including rollout number, chunk size, prefix ratios, and restarts. These factors affect probing granularity, stability, and computational cost. Results are shown in Figure~\ref{fig:opprint_ablation}.

\noindent\textbf{Effect of rollout number and chunk size. }Figure~\ref{fig:ablation_graybox_heatmap} shows that performance improves as the rollout number increases to 32, after which gains become marginal. A chunk size of 64 achieves the best or near-best results, balancing local noise and over-smoothing. OP-Print-G-Max consistently outperforms OP-Print-G, indicating that membership signals are concentrated at highly informative discrepancy positions.

\noindent\textbf{Effect of prefix-ratio selection. }Figure~\ref{fig:ablation_blackbox_prefix} shows that using more prefix ratios improves AUC and TPR@5\%FPR by probing the generation trajectory more densely. However, the gains are gradual, and the default setting already performs strongly. This suggests that the black-box signal is not limited to a narrow truncation point and remains robust under practical prefix selection.

\noindent\textbf{Effect of restart count. }Figure~\ref{fig:ablation_restart_tradeoff} shows that increasing the number of restarts improves the performance, but with diminishing returns and substantially higher cost. This suggests that additional restarts mainly reduce estimation variance rather than reveal new information. The default setting provides a favorable cost-performance trade-off.

\section{Conclusion}
We study membership inference risks in on-policy self-distillation and identify a sparse, structured correction imprint concentrated on uncertainty-sensitive tokens. Based on this observation, we propose \textbf{OP-Print}, a unified gray-box and black-box attack framework that captures either privileged teacher--student discrepancies or their observable restart-based behavioral traces. Experiments across models and distillation schemes show that OP-Print consistently outperforms existing works.

\section*{Limitaions}
The black-box attack requires repeated restart generation, which introduces additional inference cost (but the adversaries can select the attack configuration themselves). We focus on attack effectiveness and leave defense design against correction-imprint leakage to future work.

\bibliography{custom}

\newpage
\appendix

\section{Training Details}
\label{train_detail}

\paragraph{Training data.}
All target models are trained on the mathematical reasoning subset of OpenThoughts~\cite{guha2025openthoughts}.
Following the standard protocol used in our evaluation, we first randomly partition the original training set into two disjoint halves using a fixed random seed (i.e., 42), denoted as the \textit{train} split and the \textit{held-out} split. 
We then uniformly sample 500 examples from each split to construct the final member and non-member pools used throughout the paper. 
The member subset is used for target-model training, while the held-out subset is reserved for non-member evaluation only.

\begin{table*}[h]
\centering
\small
\caption{Training configurations for target models. All models are trained under the same on-policy self-distillation pipeline; only model scale and a few runtime settings differ across runs.}
\label{tab:training_details_opsd}
\begin{tabular}{lccc}
\toprule
Configuration & Qwen3-1.7B & Qwen3-4B & Qwen3-8B \\
\midrule
Number of epochs & 3 & 3 & 3 \\
Learning rate & $5\times 10^{-6}$ & $5\times 10^{-6}$ & $5\times 10^{-6}$ \\
Precision & bfloat16 & bfloat16 & bfloat16 \\
Attention implementation & SDPA & SDPA & SDPA \\
Gradient checkpointing & Yes & Yes & Yes \\
Max completion length & 1024 & 1024 & 1024 \\
Max sequence length & 20000 & 20000 & 20000 \\
Temperature & 1.1 & 1.1 & 1.1 \\
Top-$p$ & 0.95 & 0.95 & 0.95 \\
Top-$k$ & 20 & 20 & 20 \\
$\beta$ & 0 & 0 & 0 \\
$\lambda$ & 1 & 1 & 1 \\
vLLM GPU memory utilization & 0.4 & 0.4 & 0.4 \\
Max grad norm & 0.1 & 0.1 & 0.1 \\
JSD token clip & 0.05 & 0.05 & 0.06 \\
\bottomrule
\end{tabular}
\end{table*}
\paragraph{Teacher configuration.}
We use a fixed-teacher setup in all reported target-model training runs. 
Intuitively, the teacher corresponds to the initial policy augmented with privileged information, while the student is updated throughout training. 
This setting is used consistently across all model scales to isolate the privacy effects of on-policy self-distillation under a stable teacher signal.

\paragraph{Optimization and generation.}
The training process is conducted on 8 NVIDIA A6000 GPUs. Training uses bfloat16 precision, gradient checkpointing, and on-policy generation with maximum completion length 1024. 
Sampling uses stochastic decoding with temperature, top-$p$, and top-$k$ filtering. 
For efficient on-policy rollout generation, we use colocated vLLM during training. 
The distillation objective is computed on the generated continuation only, while prompt tokens are masked out from the loss.

\paragraph{Loss variants.}
For the standard OPSD variant, training uses a token-level generalized Jensen-Shannon divergence objective between teacher and student distributions on the sampled continuation. 
For the Tinker-style variant, the full-distribution matching loss is replaced by a sampled-token policy-improvement objective based on the teacher-student log-probability difference on the same rollout. 
In both cases, the teacher signal is applied on the student's own on-policy generations.

\paragraph{Implementation notes.}
All runs use the same random split seed for data construction. 
We save checkpoints periodically during training and use the resulting checkpoints in downstream membership inference evaluation. 
Unless explicitly stated otherwise, all attack results are reported on checkpoints trained under the settings summarized in Table~\ref{tab:training_details_opsd}.

\section{Additional Analysis of the Black-box Score}
\label{app:blackbox_theory}

In this appendix, we provide additional intuition and lightweight theoretical support for the black-box component of OP-Print. 
Our goal is not to claim a fully general identifiability theorem, but to formalize why restart-based output stability is a reasonable surrogate for the hidden correction imprint, why repeated restarts reduce estimation variance, and why a more concentrated output basin leads to a larger black-box score.

\subsection{Restart-text Stability as a Surrogate for the Hidden Correction Imprint}
\label{app:surrogate}

Recall that in the main text, the key challenge of the black-box setting is that the adversary cannot directly observe the privileged teacher-student discrepancy along the rollout.
We therefore introduce a latent variable that summarizes how strongly the correction process has already been internalized.

\noindent\textbf{Latent correction strength.}
For an input example $x=(q,r)$ and a prefix ratio $\rho$, let
\(
C(x,\rho)\in\mathbb{R}
\)
denote the \emph{latent correction strength} at that prefix.
Intuitively, a larger $C(x,\rho)$ means that by the time the model reaches the prefix $\pi_\rho(y)$, the privileged correction relevant to $x$ has already been more strongly absorbed into the model's local decision geometry. We do not assume that $C(x,\rho)$ is directly observable.
Instead, we assume that the distribution of restarted continuations from the prefix
\(
\pi_\rho(y^{(m)})
\)
depends on $C(x,\rho)$ through its effect on local basin concentration.

\noindent\textbf{Population-level restart stability.}
Let $\tilde Y$ and $\tilde Y'$ be two independent restarted continuations sampled from the same prefix-conditioned restart distribution at ratio $\rho$.
Define the population restart-text stability as
\begin{equation}
\label{eq:population_stability}
\mathrm{Stab}_{\mathrm{text}}^\star(x,\rho)
:=
\mathbb{E}\!\left[
\mathrm{sim}(\tilde Y,\tilde Y')
\,\middle|\,
x,\rho
\right].
\end{equation}
This quantity measures the expected pairwise similarity between two independent restarts from the same truncated state.

\noindent\textbf{Monotone surrogate assumption.}
The central assumption behind our black-box attack is that this population stability is monotone in the latent correction strength:
\begin{equation}
\label{eq:monotone_surrogate}
\mathrm{Stab}_{\mathrm{text}}^\star(x,\rho)
=
g\!\bigl(C(x,\rho)\bigr),
\end{equation}
where $g(\cdot)$ is non-decreasing. This assumption is natural: if the hidden correction has been more strongly internalized, then restarted decoding from an intermediate prefix should be more likely to remain in the same local solution basin, producing more stable continuations and hence higher pairwise text similarity.

\noindent\textbf{From prefix-level surrogate to example-level surrogate.}
Suppose that the adversary uses a set of prefix ratios $\mathcal{P}\subset(0,1)$.
Then the population black-box score can be written as
\begin{equation}
\label{eq:population_black_score}
S_{\mathrm{black}}^\star(x)
:=
\frac{1}{|\mathcal{P}|}
\sum_{\rho\in\mathcal{P}}
\mathrm{Stab}_{\mathrm{text}}^\star(x,\rho).
\end{equation}
Substituting Equation~\eqref{eq:monotone_surrogate} into Equation~\eqref{eq:population_black_score} gives
\begin{equation}
\label{eq:population_black_score_g}
S_{\mathrm{black}}^\star(x)
=
\frac{1}{|\mathcal{P}|}
\sum_{\rho\in\mathcal{P}}
g\!\bigl(C(x,\rho)\bigr).
\end{equation}
Therefore, the population black-box score is an average monotone transform of the latent correction strength across prefix ratios. Now suppose that member examples tend to exhibit larger latent correction strength on average, i.e.,
\begin{equation}
\label{eq:member_correction_domination}
\begin{aligned}
\mathbb{E}\!\left[C(x,\rho)\mid x\in\mathcal{D}_{\mathrm{mem}}\right]
&>
\mathbb{E}\!\left[C(x,\rho)\mid x\in\mathcal{D}_{\mathrm{non}}\right] \\
&\hspace{2.5em}
\text{for many } \rho\in\mathcal{P}.
\end{aligned}
\end{equation}
Because $g$ is non-decreasing, Equation~\eqref{eq:population_black_score_g} implies that
\begin{equation}
\label{eq:member_score_domination}
\begin{aligned}
\mathbb{E}\!\left[
S_{\mathrm{black}}^\star(x)
\mid x\in\mathcal{D}_{\mathrm{mem}}
\right]
&> \\
\mathbb{E}\!\left[
S_{\mathrm{black}}^\star(x)
\mid x\in\mathcal{D}_{\mathrm{non}}
\right].
\end{aligned}
\end{equation}
Thus, under this weak monotonicity assumption, restart-text stability acts as an order-preserving surrogate for membership evidence in expectation.

\noindent\textbf{Connection to the empirical score.}
In practice, we do not observe the population score in Equation~\eqref{eq:population_black_score}; instead, we estimate it with finitely many full generations and finitely many restarts:
\begin{equation}
\label{eq:empirical_black_score_appendix}
S_{\mathrm{black}}(x)
=
\frac{1}{M|\mathcal{P}|}
\sum_{m=1}^M
\sum_{\rho\in\mathcal{P}}
\widehat{\mathrm{Stab}}_{\mathrm{text}}\!\left(\mathcal{R}_{m,\rho}(x)\right).
\end{equation}
Hence, the black-box attack can be viewed as a Monte Carlo estimator of a population-level surrogate for the hidden correction imprint.

\subsection{Why Increasing the Number of Restarts Reduces Estimation Variance}
\label{app:variance_reduction}

We now formalize why increasing the restart count $R$ stabilizes the black-box estimate.

\noindent\textbf{Pairwise estimator.}
Fix an example $x$, a sampled full generation index $m$, and a prefix ratio $\rho$.
Let
\[
\tilde Y_1,\dots,\tilde Y_R
\]
be $R$ restarted continuations sampled independently from the same prefix-conditioned restart distribution.
The empirical restart-text stability is
\begin{equation}
\label{eq:u_statistic_estimator}
\widehat{\mathrm{Stab}}_{\mathrm{text}}
=
\frac{2}{R(R-1)}
\sum_{1\le i<j\le R}
\mathrm{sim}(\tilde Y_i,\tilde Y_j).
\end{equation}

\noindent\textbf{Step 1: Unbiasedness.}
Define the kernel
\(
h(\tilde Y_i,\tilde Y_j)
=
\mathrm{sim}(\tilde Y_i,\tilde Y_j).
\)
Because all unordered pairs $(\tilde Y_i,\tilde Y_j)$ are identically distributed for $i\neq j$, we have
\(
\mathbb{E}[h(\tilde Y_i,\tilde Y_j)]
=
\mathbb{E}\!\left[
\mathrm{sim}(\tilde Y,\tilde Y')
\right]
=
\mathrm{Stab}_{\mathrm{text}}^\star(x,\rho),
\)
where $\tilde Y,\tilde Y'$ are i.i.d.\ draws from the same restart distribution.
Taking expectation on both sides of Equation~\eqref{eq:u_statistic_estimator},
\begin{align}
\mathbb{E}\!\left[\widehat{\mathrm{Stab}}_{\mathrm{text}}\right]
&=
\frac{2}{R(R-1)}
\sum_{1\le i<j\le R}
\mathbb{E}[h(\tilde Y_i,\tilde Y_j)]
\nonumber\\
&=
\frac{2}{R(R-1)}
\binom{R}{2}
\mathrm{Stab}_{\mathrm{text}}^\star(x,\rho)
\nonumber\\
&=
\mathrm{Stab}_{\mathrm{text}}^\star(x,\rho).
\label{eq:unbiasedness}
\end{align}
So the empirical pairwise stability is an unbiased estimator of the population stability.

\noindent\textbf{Step 2: Variance scaling.}
Equation~\eqref{eq:u_statistic_estimator} is a second-order U-statistic.
Standard U-statistic theory implies that, under mild boundedness assumptions on the kernel $h$,
\begin{equation}
\label{eq:u_statistic_variance}
\mathrm{Var}\!\left(\widehat{\mathrm{Stab}}_{\mathrm{text}}\right)
=
O\!\left(\frac{1}{R}\right).
\end{equation}
The intuition is simple: although there are $\binom{R}{2}$ pairs, these pairs are not fully independent because they share restart samples; as a result, the variance decays more slowly than $O(R^{-2})$ but still decreases at rate $O(R^{-1})$.

\noindent\textbf{Step 3: Effect on the final black-box score.}
The final black-box score averages these estimators across multiple sampled full generations and prefix ratios:
\begin{equation}
S_{\mathrm{black}}(x)
=
\frac{1}{M|\mathcal{P}|}
\sum_{m=1}^M
\sum_{\rho\in\mathcal{P}}
\widehat{\mathrm{Stab}}_{\mathrm{text}}\!\left(\mathcal{R}_{m,\rho}(x)\right).
\end{equation}
If the restart estimators across $(m,\rho)$ pairs are weakly dependent or approximately independent, then averaging further reduces variance roughly by a factor of $M|\mathcal{P}|$.
Hence:
\begin{itemize}
    \item increasing $R$ reduces the variance of each prefix-level stability estimate;
    \item increasing $M$ or $|\mathcal{P}|$ further stabilizes the final score through averaging.
\end{itemize}
This provides a simple theoretical explanation for the empirical trend observed in the ablation study: more restarts improve stability, but with diminishing returns.

\subsection{A Simple Two-State Toy Example}
\label{app:toy_basin}

We finally give a toy model that explains why a more concentrated output basin leads to a larger black-box score.

\noindent\textbf{Two-basin assumption.}
Suppose that for a fixed prefix, restarted decoding falls into one of two local basins:
\begin{itemize}
    \item a dominant basin $A$, with probability $p$;
    \item an alternative basin $B$, with probability $1-p$.
\end{itemize}
Assume that texts generated within the same basin are more similar than texts generated across different basins:
\begin{align}
\mathbb{E}[\mathrm{sim}(\tilde Y_i,\tilde Y_j)\mid \tilde Y_i,\tilde Y_j\in A] &= s_{\mathrm{in}},
\nonumber\\
\mathbb{E}[\mathrm{sim}(\tilde Y_i,\tilde Y_j)\mid \tilde Y_i,\tilde Y_j\in B] &= s_{\mathrm{in}},
\nonumber\\
\mathbb{E}[\mathrm{sim}(\tilde Y_i,\tilde Y_j)\mid \tilde Y_i\in A,\tilde Y_j\in B] &= s_{\mathrm{out}},
\label{eq:toy_similarity_levels}
\end{align}
with
\[
s_{\mathrm{in}} > s_{\mathrm{out}}.
\]

\noindent\textbf{Step 1: Enumerate pair types.}
For two independent restarted outputs, there are three cases:
\begin{itemize}
    \item both are in $A$, with probability $p^2$;
    \item both are in $B$, with probability $(1-p)^2$;
    \item one is in $A$ and the other in $B$, with probability $2p(1-p)$.
\end{itemize}

\noindent\textbf{Step 2: Compute expected pairwise similarity.}
Using the law of total expectation,
\begin{align}
\mathbb{E}[\mathrm{sim}(\tilde Y_i,\tilde Y_j)]
&=
\bigl[p^2+(1-p)^2\bigr]s_{\mathrm{in}}
\nonumber\\
&\quad+
2p(1-p)s_{\mathrm{out}}.
\label{eq:toy_raw}
\end{align}
Now use the identity
\(
p^2+(1-p)^2 = 1-2p(1-p),
\)
to rewrite Equation~\eqref{eq:toy_raw} as
\begin{align}
\mathbb{E}[\mathrm{sim}(\tilde Y_i,\tilde Y_j)]
&=
\bigl(1-2p(1-p)\bigr)s_{\mathrm{in}}
\nonumber\\
&\quad+
2p(1-p)s_{\mathrm{out}}
\nonumber\\
&=
s_{\mathrm{in}}
-
2p(1-p)(s_{\mathrm{in}}-s_{\mathrm{out}}).
\label{eq:toy_rewrite_1}
\end{align}
Equivalently,
\begin{equation}
\label{eq:toy_rewrite_2}
\mathbb{E}[\mathrm{sim}(\tilde Y_i,\tilde Y_j)]
=
s_{\mathrm{out}}
+
\bigl(p^2+(1-p)^2\bigr)(s_{\mathrm{in}}-s_{\mathrm{out}}).
\end{equation}

\noindent\textbf{Step 3: Interpret the shape.}
The quantity
\(
p^2+(1-p)^2
\)
is minimized at $p=\tfrac12$ and increases as $p$ moves toward $0$ or $1$.
Therefore, the expected pairwise similarity is smallest when restart trajectories are evenly split across the two basins, and largest when nearly all restarts collapse into one basin.

Since $s_{\mathrm{in}} > s_{\mathrm{out}}$, Equation~\eqref{eq:toy_rewrite_2} shows that:
\begin{itemize}
    \item a \emph{dispersive} output basin (mass spread across multiple competing continuations) yields lower restart-text stability;
    \item a \emph{concentrated} output basin (mass concentrated on one dominant continuation mode) yields higher restart-text stability.
\end{itemize}

\noindent\textbf{Step 4: Connection to membership inference.}
Suppose that member examples tend to have a more strongly internalized correction process, which makes their prefix-conditioned restart distribution more concentrated in one dominant basin.
Then their corresponding $p$ values are typically closer to $0$ or $1$, whereas non-member examples are more likely to have $p$ values closer to $\tfrac12$.
By Equation~\eqref{eq:toy_rewrite_2}, this implies
\[
\begin{aligned}
\mathbb{E}[\mathrm{Stab}_{\mathrm{text}}]
&\quad\text{is larger for members} \\
&\quad\text{than for non-members.}
\end{aligned}
\]
This toy model therefore provides a simple interpretation of the black-box score: members are easier to attack because hidden correction makes their restart trajectories collapse into a tighter output basin.

\section{Extended Related Work}
\label{app:extended-related-work}

\paragraph{On-Policy Distillation and Self-Distillation.}
Knowledge distillation transfers supervision from a teacher to a student, while on-policy distillation reduces training-inference mismatch by learning from student-generated trajectories~\citep{oprel_agarwal2024}. Recent self-distillation approaches further eliminate external teachers by exploiting privileged contexts or reference solutions~\citep{oprel_zhao2026}. Related research has explored reliable supervision through semantic label verification~\citep{zhang2025trustclip}, uncertainty-guided expert selection~\citep{zhang2025choose}, and generalization-preserving continual learning~\citep{zhang2025generalization,zhang2026rep}. Despite their different objectives, these approaches highlight how training signals influence learned representations and subsequent model behavior. However, the privacy implications of privileged-context self-distillation remain insufficiently understood. OP-Print investigates whether the coupled teacher-student conditioning process leaves distinctive membership-related imprints.

\paragraph{Membership Inference and Localized Information Leakage.}
Membership inference attacks (MIAs) determine whether a sample was used during model training~\citep{oprel_shokri2017,oprel_carlini2022}. Existing LLM attacks exploit token probabilities, sequence likelihoods, and calibrated uncertainty statistics~\citep{oprel_zhang2024minkpp}, while recent studies emphasize the importance of controlling distributional confounders~\citep{oprel_meeus2025,oprel_puerto2025}. The privacy risks of knowledge distillation have also attracted attention~\citep{oprel_jagielski2023,oprel_zhang2025kd}. Beyond privacy research, selective information modeling has been explored through cross-spectral augmentation~\citep{lei2024rle}, modality-decoupled representations~\citep{feng2025mdreid}, sparse mixture-of-experts~\citep{feng2025multi}, saliency-guided feature transfer~\citep{tan2024occluded}, attention-based alignment~\citep{tan2026knowing}, and dynamic cross-modal coordination~\citep{lin2026dyco}. While addressing different problems, these studies illustrate the importance of identifying informative components rather than treating all information uniformly. OP-Print similarly examines localized training signals, revealing that membership information concentrates on uncertainty-sensitive correction sites rather than being uniformly distributed across generated trajectories.

\paragraph{Black-Box Behavioral Signals and Structured Dependencies.}
When internal model statistics are unavailable, membership inference must rely on observable generation behavior. Existing approaches investigate output consistency and memorization patterns, while privacy auditing further examines how training data influence model responses~\citep{oprel_panda2025}. Related challenges concerning structured information dependencies and output reliability arise in multimodal hallucination mitigation~\citep{shu2026semantics}, dense temporal correspondence modeling~\citep{wu2026pairformer}, and constrained spatiotemporal generation~\citep{wu2025animateqr}. OP-Print focuses on autoregressive generation, where intermediate prefixes constrain subsequent continuations. Its black-box attack measures restart-based output stability as an observable surrogate for hidden correction imprints, complementing the gray-box attack that directly aggregates privileged teacher-student discrepancies. Together, these approaches characterize membership leakage under different levels of model access.

\end{document}